%% file: root.tex
\documentclass[letterpaper, 10 pt, conference]{ieeeconf}  

\IEEEoverridecommandlockouts                              

\usepackage{latexsym}
\usepackage{times}

\usepackage{multicol}

\usepackage{array}

\usepackage[caption=false,font=normalsize,labelfont=sf,textfont=sf]{subfig}
\usepackage{textcomp}
\usepackage{stfloats}
\usepackage{url}
\usepackage{verbatim}
\usepackage{graphicx}

\usepackage{anyfontsize}
\usepackage{float}
\usepackage{caption}
\usepackage{url}            
\usepackage{booktabs}       
\usepackage{amsmath,amsfonts}       
\usepackage{nicefrac}       
\usepackage{microtype}      
\usepackage[dvipsnames]{xcolor}
\usepackage[ruled,vlined,linesnumbered]{algorithm2e}
\let\labelindent\relax
\usepackage{enumitem}
\usepackage{bm}
\usepackage{multirow}
\usepackage{threeparttable}
\usepackage{adjustbox}
\usepackage{eqparbox}
\usepackage{arydshln}
\usepackage{wrapfig}

\usepackage{graphicx}
\usepackage{svg}
\usepackage{amsmath,amssymb} 
\usepackage{tabularx}
\usepackage{multirow}
\usepackage{bbm}
\usepackage{dsfont}
\usepackage{cuted}
\usepackage{flushend}
\usepackage{lipsum}

\usepackage[bookmarks=true]{hyperref}
\usepackage{threeparttable}
\usepackage{hyperref}
\hypersetup{
    colorlinks,
    linkcolor={red!50!black},
    citecolor={blue!50!black},
    urlcolor={blue!80!black}
}

\NoCaptionOfAlgo

\title{
Deep Bayesian Future Fusion for \\
Self-Supervised, High-Resolution, Off-Road Mapping
\vspace{-0.4cm}
}

\author{Shubhra Aich$^{1}$, Wenshan Wang$^{1}$, Parv Maheshwari$^{1}$, Matthew Sivaprakasam$^{1}$, 
Samuel Triest$^{1}$, Cherie Ho$^{1}$,\\ 
Jason M. Gregory$^{2}$, John G. Rogers III$^{2}$, and Sebastian Scherer$^{1}$
\vspace{0.2cm}
\\ 
$^{1}$\texttt{Robotics Institute, Carnegie Mellon University, Pittsburgh, PA, USA}\\ 
$^{2}$\texttt{DEVCOM Army Research Laboratory, Adelphi, MD, USA}
}

\makeatletter
\newcommand{\removelatexerror}{\let\@latex@error\@gobble}
\makeatother

\begin{document}

\title{\LARGE \bf
Deep Bayesian Future Fusion for \\
Self-Supervised, High-Resolution, Off-Road Mapping
\vspace{-0.2cm}
}







\author{Shubhra Aich$^{1}$, Wenshan Wang$^{1}$, Parv Maheshwari$^{1}$, Matthew Sivaprakasam$^{1}$, 
Samuel Triest$^{1}$, Cherie Ho$^{1}$,\\ 
Jason M. Gregory$^{2}$, John G. Rogers III$^{2}$, and Sebastian Scherer$^{1}$ 
\thanks{* This work was supported by ARL awards \#W911NF1820218 and \#W911NF20S0005.}%
\thanks{$^{1}$ Robotics Institute, Carnegie Mellon University, Pittsburgh, PA, USA. \{saich,wenshanw,parvm,msivapra,striest,cherieh,basti\}@andrew.cmu.edu}%
\thanks{$^{2}$ DEVCOM Army Research Laboratory, Adelphi, MD, USA. \{jason.m.gregory1,john.g.rogers59\}civ@army.mil}%
}


\maketitle
\begin{strip}
  \centering
  \vspace*{-27mm}
  \includegraphics[width=\linewidth]{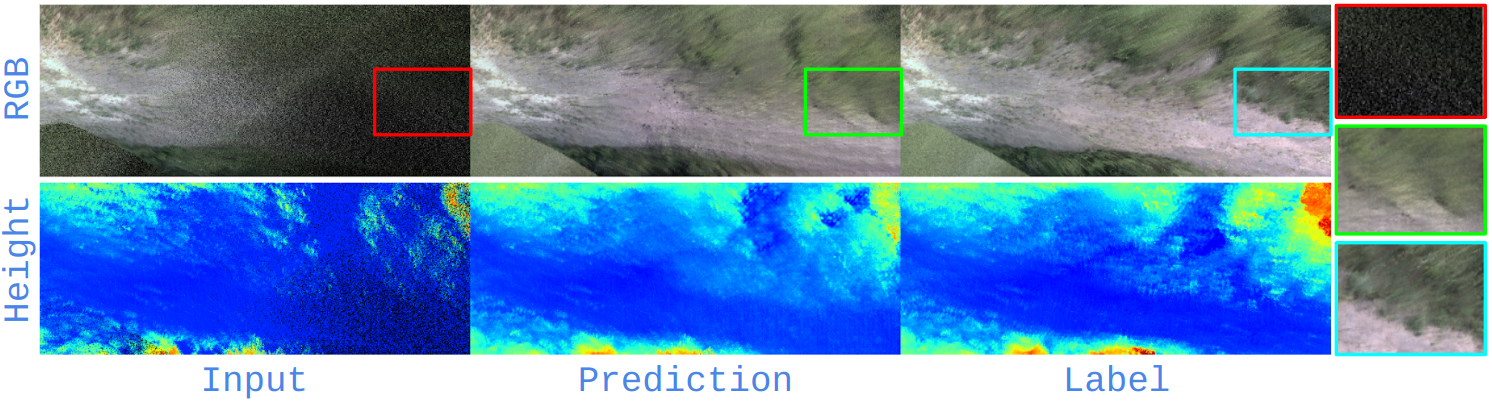}
  \captionof{figure}{RGB (top) and height (bottom) Bird's eye view (BEV) maps (cropped at $12m \times 30m$) with $2cm$ pixel resolution from the sparse input (left),
future-fusion label (right), and prediction from our Bayesian UNet/e2 structure trained with these input/label 
pairs (middle column). Best viewed in digital format. The images are downsampled to comply with the size limit.}
  \label{fig:title}
  \vspace*{-1mm}
\end{strip}

\IEEEpeerreviewmaketitle

\noindent \begin{abstract}
High-speed off-road navigation requires long-range, high-resolution maps to enable robots to safely navigate over different surfaces while avoiding dangerous obstacles. However, due to limited computational power and sensing noise, most approaches to off-road mapping focus on producing coarse (20-40cm) maps of the environment. In this paper, we propose Future Fusion, a framework capable of generating dense, high-resolution maps from sparse sensing data (30m forward at 2cm). This is accomplished by -- (1) the efficient realization of the well-known Bayes filtering within the standard deep learning models that explicitly accounts for the sparsity pattern in stereo and LiDAR depth data, and (2) leveraging perceptual losses common in generative image completion. The proposed methodology outperforms the conventional baselines. Moreover, the learned features and the completed dense maps lead to improvements in the downstream navigation task.
\end{abstract}


\IEEEpeerreviewmaketitle

\input sections/introduction.tex

\input sections/literature.tex
\input sections/method.tex

\input sections/experiments.tex
\input sections/conclusion.tex









\bibliographystyle{IEEEtran}
\bibliography{
    bibliographies/main, 
    bibliographies/inpainting,
    bibliographies/gan,
    bibliographies/army,
    bibliographies/datasets,
    bibliographies/sdc,
    bibliographies/fusion,
    bibliographies/forecasting,
    bibliographies/depth_completion,
    bibliographies/theory,
    bibliographies/misc,
    bibliographies/offroad
}

\end{document}

%% file: sections/introduction.tex
\section{Introduction}
\label{sec:intro}


\begin{figure}[!hb]
\vspace{-0.5cm}
\includegraphics[width=0.48\textwidth]{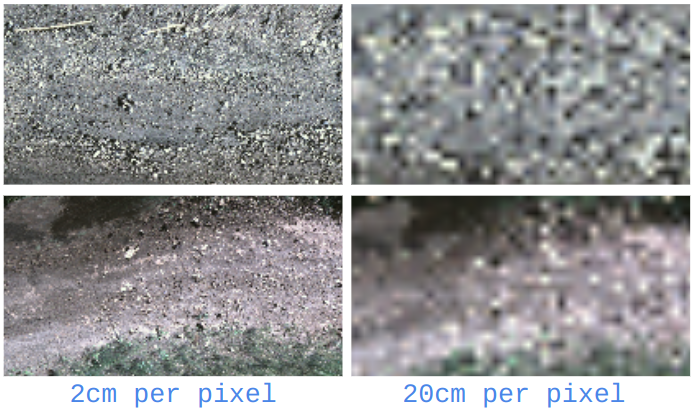}
\vspace{-0.1cm}
\caption{
Sample BEV images (cropped) showing the same region with different pixel resolutions -- $2$cm used in this work and $20$cm which is the finest in recent literature \cite{roadrunner}. The difference in the perceptual quality for a typical off-road terrain is evident.
}
\vspace{-0.5cm}
\label{fig:motivation}
\end{figure}


Off-road environments are challenging for robots due to the existence of many small obstacles (e.g. thin ditches, rocks, etc.) and varied surface types. In order for a robot to safely traverse these environments at high speed, it must be able to accurately map its environment at a high resolution. Most off-road perception stacks create a map with coarse resolution (20cm-40cm \cite{roadrunner, terrain-net}). This is due to both memory constraints, and data sparsity at long range. However, as shown in Figure \ref{fig:motivation}, this coarse resolution removes significant detail in the local terrain, preventing downstream planning and control from reasoning about the different surface types.
Simply increasing resolution is a non-trivial problem, as data becomes quadratically sparser as resolution increases. Furthermore, the problem of predicting unobserved terrain becomes more challenging as distance to the robot increases.
 
In this work, we train a network to predict dense BEV maps given the sparse BEV map as input (Figure \ref{fig:title}). In contrast to prior work, we focus on this task at a very high resolution of 2cm, enabling accurate mapping of complex surface types such as gravel and vegetation. Additionally, we propose a novel network architecture and training method that enables efficient high-resolution, geometrically-consistent inpainting at long range. In contrast to existing work \cite{roadrunner, terrain-net}, which elects to produce traversability and semantics, we elect to produce raw RGB and height maps. This enables our method to integrate cleanly into existing downstream traversability methods.

In order to train a network for high-resolution map completion, we create a dataset of input/label pairs from both stereo RGB and LiDAR point clouds, along with an efficient CPU/GPU-based local mapper. We then train a novel Bayes filter-inspired network \cite{prbook} with SoTA methods in visual generative modeling \cite{stylegan, stylegan2-ada} to predict (i.e. complete and refine) BEV maps with high-fidelity (Figure \ref{fig:title}, \ref{fig:sample-results}).
In addition, we demonstrate the efficacy of both the predicted maps and the associated deep features on the downstream task of 
costmap prediction from highly sparse proprioceptive sensing data.

\noindent In summary, our contributions are as follows:

\begin{itemize}[leftmargin=*]


\item We propose a scalable data-generation protocol for self-supervised, dense map completion. With that, we generate a large-scale dataset comprising high-resolution (i.e. 2 cm) (raw input / dense label) pairs for RGB and height by fusing the future map information (i.e. future-fusion). 

\item 
We refine and complete the input BEV maps with an efficient
convolutional-recurrent mechanism, which is a realization of the well-known Bayes filter \cite{prbook}. We demonstrate that this recurrent method outperforms the conventional counterparts that predict the map in a single shot. We call this framework \textbf{Deep Bayesian Future-Fusion (DBFF)}.

\item We validate the DBFF map and learned features
on the downstream task of costmap estimation for off-road navigation, 
demonstrating improved performance.

\item We qualitatively show that the DBFF encoder features contain distinctive information 
about terrain categories and noise and sparsity in the maps learned purely in a self-supervised manner.

\end{itemize}

%% file: sections/literature.tex
\section{Related Works}
\label{sec:literature}

\noindent \textbf{Irregular hole-filling and image inpainting: }
Recent works on image inpainting and depth completion address the issue of
irregular shaped holes with different forms of the standard convolution operation \cite{gated-convolution, partial-convolution, sparsity-invariant-cnn}. However, unlike plain convolutions, 
these customized convolution operations equipped with various normalization tricks are quite difficult to accelerate via matrix multiplication \cite{caffe-phd-thesis}. We have empirically found that these approaches do not directly improve performance on our task.
CRA \cite{ultra-high-res} proposes a low-resolution hole-filling followed by 
adversarial low to high resolution transfer. For our task, where the input comprises salt and pepper like holes, the pooled low resolution images destroy the texture distributions of the images and thus take the low resolution transfer task severely out-of-distribution, which we empirically find difficult to recover in the later stage. 

There exists additional work that leverages a combination of transformers and diffusion models \cite{ddim, ddpm, dit, vit}. However, such methods exceed the computational limits of what most off-road robots \cite{tartandrive} can support in real time.
\vspace{0.2cm}

\noindent \textbf{Point cloud forecasting: }
The closest subdomain to our BEV map completion task for offroad driving is point cloud forecasting \cite{forecasting-st3dcnn, forecasting-spfnet, stochastic-spfnet, occ-forecasting-4d} where the points at future time steps are predicted from the previous scans. 
The spatio-temporal 3D CNN \cite{forecasting-st3dcnn} 
predicts the future projections in 2D spherical view by stacking the scans from the past. SPF \cite{forecasting-spfnet} utilizes a similar forecasting pipeline for detection and tracking afterward. 
Stochastic SPF \cite{stochastic-spfnet} incorporates probabilistic generative modeling into SPF via temporally-conditioned latent space learning. 
The 4D occupancy network \cite{occ-forecasting-4d} bypasses the exact points prediction problem with an easier proxy of 4D occupancy grid forecasting. However, these methods are typically focused on occupancy and object prediction, which are not necessarily suitable to off-road domains.
\vspace{0.2cm}

\noindent \textbf{Image-guided monocular depth completion: }
Another subdomain similar to our BEV RGB/height completion task is 
supervised monocular depth completion (MDC)
\cite{kitti-depth-completion-leaderboard}. Recent works on supervised MDC focus on the design of better multimodal
fusion networks from FPV \cite{mffnet} and BEV \cite{bev-dc}, 
gradual multi-stage refinement \cite{rignet}, 
learnable attention-based propagation of spatial affinity among neighboring pixels \cite{dy-spn}, and amalgamation of convolution and SOTA transformer modules \cite{completion-former}.
Nonetheless, supervised MDC is arguably an easier problem for a couple of reasons. 
from sparse depth input. 
There exist stronger priors for prediting depth in FPV space
\cite{how-nn-see-depth, vis-nn-mde, banet-mde}. Additionally, networks for our task must also learn to de-noise inputs. 

\noindent \textbf{Inpainting in off-road navigation: }
BEVNet \cite{bev-net} predicts a BEV semantic map from on-board lidar data. TerrainNet \cite{terrain-net} further extends this, consuming stereo pairs and predicting the elevation map along with the semantic map. UNRealNet \cite{unreal-net} estimates robot-agnostic traversability features with an eye on test-time adaptation. ALTER \cite{alter} learns to approximates the dense traversability costmap in the FPV space based on geometric cues. 
Finally, RoadRunner \cite{roadrunner} learns to fuse lidar and RGB information based on the traversability pseudolabels generated by another existing navigation stack \cite{step}. 

These methods use aggregated sensing to generate pseudo-labels corresponding to the selected downstream tasks only. On the contrary, we address the general problem of input map completion and denoising through the lens of generative modeling. that enables our method to integrate into arbitrary downstream tasks.


%% file: sections/method.tex
\begin{figure*}[!htbp]
\centering
\includegraphics[width=0.9\textwidth]{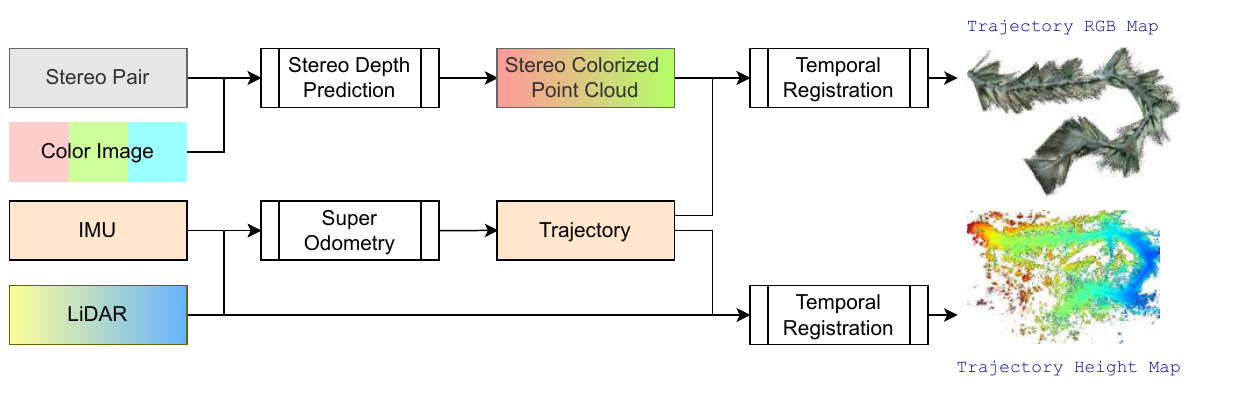}
\vspace{-0.5cm}\caption{
Block diagram of the data generation protocol via future-fusion. 
The odometry estimation provided by the Super Odometry \cite{super-odom} framework is employed to register the colorized stereo point cloud \cite{tartan-vo} and LiDAR scans to generate the trajectory RGB and height maps. 
These dense maps containing billions of points are then utilized to generate dense labels corresponding to the sparse local BEV maps for self-supervision. See Section \ref{subsec:data-gen} for details. 
}
\vspace{-0.6cm}
\label{fig:full-traj}
\end{figure*}

\section{Our Approach}
\label{sec:method}

\subsection{Data Generation by Future-Fusion}
\label{subsec:data-gen}

\noindent \textbf{Future-fusion pipeline:} As described in Section \ref{sec:intro}, the limited 
sensing range of the offroad ATVs \cite{tartandrive} leads to 
the sparse and noisy measurement starting from as close as about $6m$ from the ego-vehicle. 
This range is not sufficient to enable safe off-road planning.
Thus, to denoise and complete the map up to the serviceable range of $30m$ ahead, we must be able to predict and denoise BEV maps. In order to train a network for this task, we create a dataset of dense BEV RGB and heightmaps for $30m \times 30m$.

\begin{figure}[!htbp]
\centering
\includegraphics[width=0.49\textwidth]{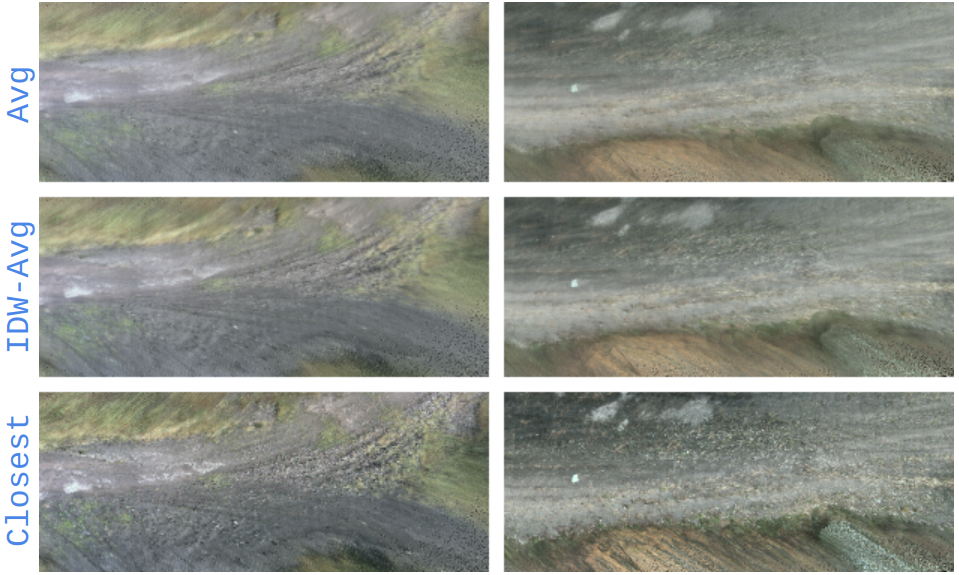} 
\vspace{-0.4cm}\caption{
Comparison of different pixel attibution strategies for BEV map generation -- unweighted average (Avg); inverse-distance weighted average (IDW-Avg); closest point attribution (Closest). The distinctive perceptual quality for the later strategy is evident. Best viewed in digital format.
}
\vspace{-0.5cm}
\label{fig:closest_point}
\end{figure}

Figure \ref{fig:full-traj} depicts the pipeline for generating our ground-truth BEV maps. The data generation protocol takes stereo pairs and monocular RGB image, LiDAR scans, and a IMU measurement as input. 
We first leverage the stereo depth prediction model from TartanVO \cite{tartan-vo} 
to provide accurate per-pixel depth. We then use Super Odometry \cite{super-odom} to provide pose estimates between depth images and transform the resuting stereo point clouds into a shared frame. Additionally, we leverage the registered LiDAR data to produce a corresponding non-colorized point cloud.


From this registered pointcloud, we can create input and label pairs by performing local mapping from the poses in the robot trajectory (in the frame of the current pose). Input maps are created using the points up to the pose's current timestep, while the label maps can use points from future timesteps. RGB maps are created from the stereo point cloud, while height maps are created using the lidar point cloud.



\noindent \textbf{Pixel attribution strategies:} 
We consider three strategies to fuse the points for each BEV pixels -- (1) (Avg) unweighted average, (2) (IDW-Avg) weighted average based on inverse distance from the ego-vehicle, and (3) (Closest) selecting the candidate with the minimum distance. Since stereo depth prediction is susceptible to significant noise, the closest candidate produces perceptually much higher quality  samples for tiny or narrow objects like gravels and mud slick (Figure \ref{fig:closest_point}). 
However, LiDAR scans are not affected by such noise within our target limit of $30m$. Therefore, we employ the inverse distance weighted averaging from above to aggregate the LiDAR height maps.
\vspace{0.1cm}

\begin{figure*}[!htbp]
\centering
\includegraphics[width=\textwidth]{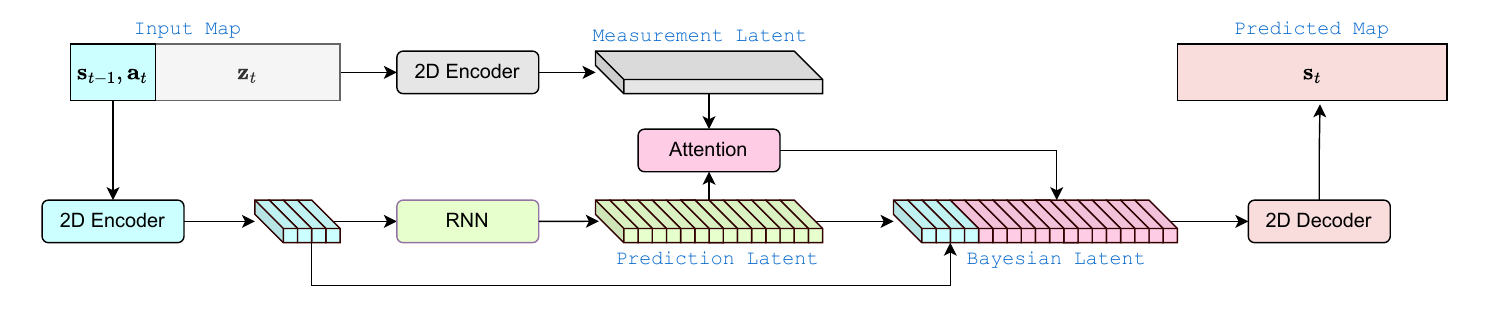}
\vspace{-0.7cm}\caption{
Schematics of the deep Bayesian fusion mechanism. The input is split into two disjoint sub-regions: (1) smaller, proximal, and reliable one resembling the combined previous state $\mathbf{s}_{t-1}$ and control action $\mathbf{a}_t$, and (2) distal, sparse and noisy one equivalent to the measurement $\mathbf{z}_t$. \textbf{(Prediction step)} The distal latent is predicted by unrolling the proximal latent via RNN. \textbf{(Measurement update)} The predicted roll out is modulated by the noisy distal latent via cross-attention mechanism. Finally, the fused latent is decoded into the complete local BEV map representing the current state $\mathbf{s}_t$. 
}
\vspace{-0.6cm}
\label{fig:bayes}
\end{figure*}

\vspace{-0.1cm}
\subsection{Deep Bayesian Future Fusion}

\noindent \textbf{Motivation:} A standard deep network trained with a combination of generative losses performs well when the input does not contain significantly varying degrees of noise or sparsity. However, we empirically find such a setup falls short for our case where such noise is present in the input as a nonlinear function of distance. Thus, to make the model explicitly aware of this noise structure, we inscribe a mechanism similar to Bayes filtering [5] in this paper.

\noindent \textbf{Preliminary background:} The governing equations underlying the problem of state estimation are given by \cite{prbook}: \vspace*{-0.5cm}

\begin{align*}
& \mathbf{s}_t = g(\mathbf{s}_{t-1}, \mathbf{a}_t) + \omega_t \\
& \mathbf{z}_t = h(\mathbf{s}_t) + \nu_t 
\end{align*}

\noindent Here, $\mathbf{s}_t, \; \mathbf{z}_t, \; \mathbf{a}_t$ denote the state representation, measurement, and control actions at time $t$. Moreover, $\omega_t$ and $\nu_t$ are the additive process and measurement noise terms, respectively. With the Markovian assumption, the posterior can be expressed as follows: \vspace*{-0.6cm}

\begin{align*}
p(\mathbf{s}_t | \mathbf{s}_{t-1}, \mathbf{a}_{t}, \mathbf{z}_{t} ) =
\eta \; p(\mathbf{z}_t | \mathbf{s}_t) \;
p(\mathbf{s}_t | \mathbf{s}_{t-1}, \mathbf{a}_{t} )
\end{align*}

\noindent In this equation, the estimation task is factorized into the prediction step (second term) followed by the measurement update (first term). In general, the approximation models act like a MAP estimator \cite{prml-bishop} for the individual factors where we can express them as follows: \vspace*{-0.5cm}

\begin{align}
\label{eq:factor}
\phi(\mathbf{s}_t | \mathbf{s}_{t-1}, \mathbf{a}_{t}, \mathbf{z}_{t} ) =
\chi \left(
\theta(\mathbf{z}_t | \mathbf{s}_t), \;
\psi(\mathbf{s}_t | \mathbf{s}_{t-1}, \mathbf{a}_{t} )
\right )
\end{align}

\noindent Here, $\theta, \; \psi$ represent the approximators (e.g. deep learning models) factorizing the underlying joint distribution of $\phi$, and $\chi$ is the composition operator used to combine the factors, i.e. prediction and noisy measurement. \\

\begin{figure}[!t]
\centering
\includegraphics[width=0.48\textwidth]{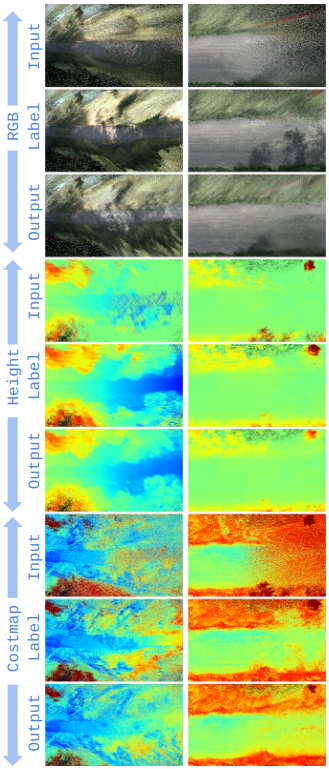}
\vspace{-0.55cm}
\caption{
Sample prediction results (cropped) for the Bayes-UNet/e2 model alongside the input and future-fusion labels for both RGB (top), height (middle), and costmap (bottom). Additional visualizations are in the supplementary video.
}
\vspace{-0.7cm}
\label{fig:sample-results}
\end{figure}

\noindent \textbf{Deep Bayesian fusion:} 
We start with a preliminary observation that the input BEV map of our servicable range ($30m \times 30m$) is quite accurate approximately in the forward (leftmost) $6m$ region (here, in the figures, left to right indicates forward movement). 
Based on that, we model the factorization in Equation 
\ref{eq:factor} below. We leverage a standard 2D encoder-decoder 
architecture, though our architecture-agnostic formulation is equally applicable to any other network structure. 

\noindent $\bullet$ \textbf{Prediction update, }$\psi(\mathbf{s}_t | \mathbf{s}_{t-1}, \mathbf{a}_{t} )$:  
The forward of $6m$ region (leftmost in the images for left to right movement) in front of the ego-vehicle
can be thought of as a composition of both the sufficient information 
about the previous state $\mathbf{s}_{t-1}$ and the control action $\mathbf{a}_t$ (i.e. tentative forward motion) for the current state $\mathbf{s}_t$. 



We realize $\psi(\mathbf{s}_t | \mathbf{s}_{t-1}, \mathbf{a}_{t} )$ 
via the 2D CNN encoder and a column-wise RNN encoder-decoder 
operating on the latent space of the former (Figure \ref{fig:bayes}).
First, we pass this initial forward (left in images) sub-region of RGB and height map 
representing our $\mathbf{s}_{t-1}$ and $\mathbf{a}_{t}$
to the 2D CNN encoder for a corresponding latent feature map.
Our RNN encoder-decoder performs horizontal (left $\rightarrow$ right)
unrolling via rolling out the latent columns one at a time. 
The RNN encoder is first initialized by the latent columns corresponding to the $6m$
sub-region proximal to the ego-vehicle. Next, the RNN decoder unrolls the latent columns for the remaining proximal to distal latent prediction. 
\vspace{0.1cm}

\noindent $\bullet$ \textbf{Measurement update, }$\chi \left(
\theta(\mathbf{z}_t | \mathbf{s}_t), \;
\hdots
\right )$ : 
As already explained above, the proximal (leftmost) $6m$ sub-region (chosen empirically)
in the forward direction 
indicates the previous state 
$\mathbf{s}_{t-1}$ and control action $\mathbf{a}_{t}$.
Thus, for our $30m$ mapping task, 
the distal $24m$ represents the sparse
measurement $\mathbf{z}_{t}$ with varying degree of noise. 
The measurement update consists of two steps. 
First, we realize $\theta(\mathbf{z}_t | \mathbf{s}_t)$ 
by encoding the $24m$ distal input $\mathbf{z}_t$ with a 
2D CNN encoder similar to the prediction update. 
Next, the measurement update $\chi$ is implemented 
by modulating the unrolled prediction from the prediction step with
this measurement latent via 
scaled dot product attention \cite{attention} as follows: 
\vspace*{-0.5cm}

\begin{align*}
\chi (\theta, \psi) = \texttt{softmax} 
\left (
\frac{ \psi \theta^{T} } {\sqrt{d_\theta}}
\right ) \theta
\end{align*}

\noindent Here, $\chi, \; \theta, \; \psi$ denotes the output of the corresponding factors from 
Equation \ref{eq:factor}, and $d_\theta$ is the embedding dimension of $\theta$. 
Finally, we employ a 2D CNN decoder to generate the dense completion of RGB and heightmap from the deep Bayesian latents $\chi (\theta, \; \psi)$ from above.
\vspace{0.1cm}

\noindent \textbf{Remarks on transformer-based fusion: } Contrary to the sequential unrolling 
mechanism provided by our recurrent structure, transformers \cite{vit, segformer}
predict the temporal dependency in a one-shot manner, and seem to outperform 
the RNN variants in recent times. However, we empirically found the SOTA 
encoder-decoder structure based on the transformer \cite{segformer, ganav} produces
blurry predictions in the distal regions.
Some potential reasons for this include the lower resolution of the prediction, 
lack of specific structure to account for additional noise on the physically distal part of the input, 
and the limited amount of training data compared to domains where transformer-based architectures have been most successful.
Overall, our bayesian fusion is also equally applicable for transformers. However,
primarily due to their lower-resolution prediction and reasonable performance of the 2D CNN,
in this paper, we proceed with the 2D CNN backbones for further analysis. 

\subsection{Loss functions}

Apart from the standard reconstruction loss 
$\mathcal{L}_{rec}$, following the recent advances in generative modeling, in particular, 
unseen view synthesis \cite{nted}, we employ adversarial $\mathcal{L}_{adv}$ and 
perceptual losses $\mathcal{L}_{perp}$
\cite{stylegan, perceptual-loss} as the additional loss functions for high-level domain alignment.
In addition, UNet is chosen as the discriminator for detailed discriminative feedback 
while training \cite{unet-discriminator}. 
Overall, the complete loss is given by 
\vspace{-0.2cm}
\begin{equation}
  \mathcal{L} = \lambda_{rec} \mathcal{L}_{rec} + \lambda_{adv} \mathcal{L}_{adv} + \lambda_{perp} \mathcal{L}_{perp}
\end{equation}
where $\lambda_{rec}$, $\lambda_{adv}$, and $\lambda_{perp}$ are the interaction hyperparameters
chosen empirically.



\begin{figure*}[!htbp]
\centering
\includegraphics[width=\textwidth]{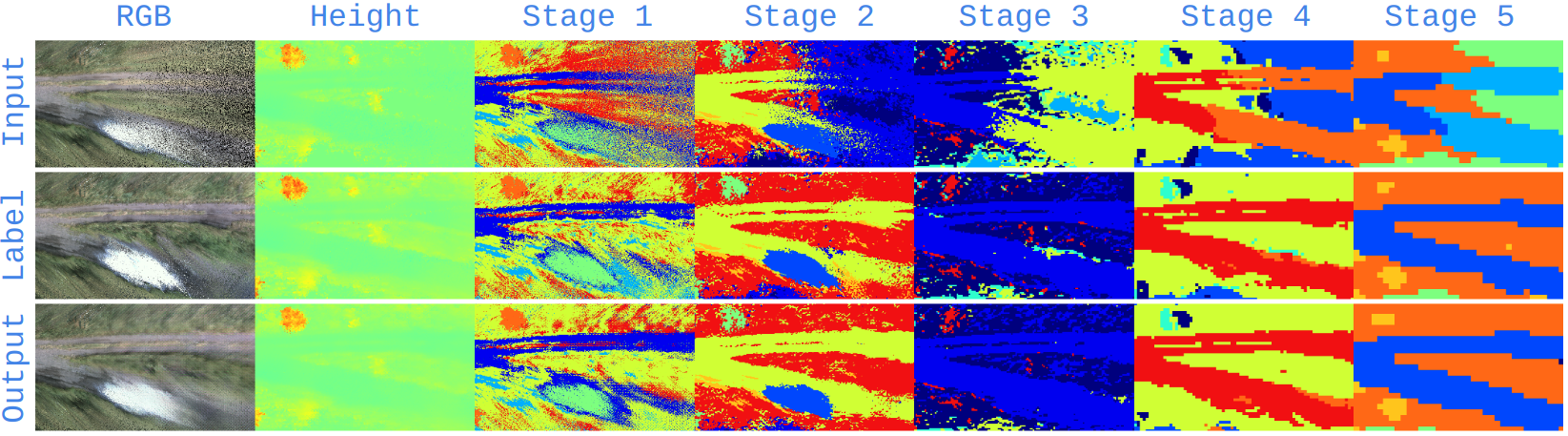}
\vspace{-0.55cm}
\caption{
Sample results for the simple k-means (k=10 chosen empirically) clustering of the stage-wise encoder features. The clusters appear from detailed to progressively more abstract towards deeper layers (Stage $1 \rightarrow 5$). More notably, the network distinguishes the regions with noisy, sparse, and somewhat corrupted inputs (RGB/height) in the deeper layers -- (top row) stage $2$ (yellow/blue), stage $3$ (blue/yellow), stage $4$ (red/orange), and stage $5$ (blue/indigo). (Middle) These noisy clusters are not present in the features extracted from dense future fusion labels. ``Output'' row shows the features extracted from the reconstructed (i.e. output) maps. The absence of noisy clusters here indicates the output features are somewhat well-aligned with the label features.
Note that the color codes for the clusters among stages are independent. Moreover, the stage-wise features are upsampled to be of the same dimension to facilitate visualization, resulting in square patterns in stages $4-5$.
}
\vspace{-0.2cm}
\label{fig:layerwise-clustering}
\end{figure*}

%% file: sections/experiments.tex
\section{Experiments}
\label{sec:experiments}

\begin{figure*}[ht]
\hspace{1cm}
\begin{minipage}[t]{9cm}%
\begin{center}
\captionof{table}{Comparison of Map Completion Models}
\label{tab:gen}
\begin{adjustbox}{width=1.25\columnwidth,center}
\renewcommand{\arraystretch}{1.3}
\begin{threeparttable}
\begin{tabular}{l|ccc|ccc}
\hline
\multicolumn{1}{c|}{\multirow{2}{*}{Method}} & \multicolumn{3}{c|}{RGB} & \multicolumn{3}{c}{Height} \\ \cline{2-7} 
 \multicolumn{1}{c|}{} & MAE ($\downarrow$) & FID ($\downarrow$) & SSIM ($\uparrow$) & MAE ($\downarrow$) & FID ($\downarrow$) & SSIM ($\uparrow$) \\ \hline
Baseline (Input) & 0.1430 & 9.1001 & 0.3551 & 0.8987 & 9.4244 & 0.5540 \\
 UNet & 0.1114 & 2.5217 & 0.3847 & 0.5943 & 1.9394 & 0.7029 \\
 UNet/e2 & 0.1106 & 1.5233 & 0.3721 & \textbf{0.3799} & 1.8139 & 0.7355 \\ 
 Bayes-UNet/e2 & \textbf{0.1082} & \textbf{1.4990} & \textbf{0.3910} & 0.4039 & \textbf{1.2323} & \textbf{0.7502} \\ 
 \hline
\end{tabular}
\begin{tablenotes}
\item[] \hspace{-0.4cm} Lower is better ($\downarrow$); \hspace{0.1cm} Higher is better ($\uparrow$)
\item []
\end{tablenotes}
\end{threeparttable}
\renewcommand{\arraystretch}{1}
\end{adjustbox}
\end{center}
\vspace{-1.05cm}
\end{minipage}
\begin{minipage}[t]{9cm}%
\begin{center}
\captionof{table}{MAE for Costmap Prediction}
\label{tab:costmap}
\begin{adjustbox}{width=0.7\columnwidth,center}
\renewcommand{\arraystretch}{1.3}
\begin{threeparttable}
\begin{tabular}{l|ccc}
\hline
\multicolumn{1}{c|}{\multirow{2}{*}{Input}} & \multicolumn{3}{c}{Speed (m/s)} \\ \cline{2-4} 
\multicolumn{1}{c|}{} & 0-4 & 4-8 & \textgreater{}8 \\ \hline
Raw (Baseline) & 0.1238 & 0.1537 & 0.1674 \\
UNet/e2 & 0.0974 & 0.1236 & 0.1227 \\
Bayes-UNet/e2 & \textbf{0.0864} & \textbf{0.0943} & \textbf{0.0986} \\ \hline
\end{tabular}
\begin{tablenotes}
\item[] \hspace{-0.4cm} Raw $\equiv$ BEV maps without network prediction
\item[] \hspace{-0.4cm} MAE $\equiv$ Mean absolute error
\item []
\end{tablenotes}
\end{threeparttable}
\renewcommand{\arraystretch}{1}
\end{adjustbox}
\end{center}
\vspace{-1.05cm}
\end{minipage}
\end{figure*}

In this section, first we describe the statistics of the generated dataset.
Next, we compare the deep Bayesian fusion mechanism against the vanilla model, 
in terms of the direct photometric loss and other metrics used for generative modeling \cite{fid-score, perceptual-loss, nted}.
We also include qualitative visualization for the clustering of our encoder features.
Next, we validate the map completion results further through the performance on the additional downstream task.

\subsection{Datasets}

The RGB and heightmap input/label pairs are generated for 28 trajectories (or runs) from the TartanDrive-2.0 \cite{tartandrive-2} dataset. The RGB and stereo pairs are from Multisense S21, and the LiDAR scans contain the points from two Velodyne VLP-32 and a Livox Mid-70 sensors. We hold out $4/28$ trajectories for testing and the rest for training. In total, there are $65134$ and $10307$ input/label BEV pairs of range $30m \times 30m$ in our train/test splits, respectively. 

\subsection{Architecture}

The recent trend of combining the diffusion mechanism with transformers \cite{ddim, ddpm, dit} demonstrates promising results for image generation. However, the runtime of the diffusion-based models are far from being real-time for the autonomy stack in the off-road ATVs. Considering that, we utilize the efficient convolutional backbone, UNet \cite{unet} with the recent advances in GAN-based loss formulations \cite{stylegan, perceptual-loss} in this paper. For the $30m \times 30m$ map completion, the runtime is $16$Hz on a \texttt{NVIDIA V100 GPU}.  

\subsection{Photometric and Generative Evaluation}

\noindent \textbf{Comparison to off-road benchmarks:} As noted in related works, existing off-road literature \cite{alter, roadrunner, unreal-net, step, bev-net, terrain-net} directly optimize the 2D FPV image features for a set of downstream tasks and potentially uplift to 3D. Contrarily, we aim for a generic, high-resolution, BEV map completion to be used for arbitrary tasks down the line. Thus, to our knowledge, there are no off-road benchmarks available for a direct comparison. Instead, we compare against the conventional network structures for benchmarking next.

\noindent \textbf{Comparison to conventional baselines:} Table \ref{tab:gen} shows the comparative results for generative modeling. The raw input is our baseline that also serves as the lower bound of performance. For UNet, we stack the RGB and height channels altogether as input, which acts like an early fusion mechanism. For UNet/e2, we split the UNet encoder into half and process RGB and height independently, and merge in the decoding step. This formation resembles a late fusion mechanism. 
Since UNet/e2 outperforms UNet, we implement our Bayesian fusion mechanism on top of UNet/e2, which we call Bayes-UNet/e2. 
We find that our method's runtime is competitive with existing architectures. The Bayesian variant outperforms its vanilla counterparts in most cases (Table \ref{tab:gen}).

\subsection{Stage-Wise Feature Clustering}
\label{subsec:cluster}

We further qualitatively investigate the types of features learned through our self-supervised denoising and completion task, particularly in the off-road BEV setting. To our knowledge, such qualitative analysis is the first of its kind for self-supervised off-road perception. 

More specifically, we take the features from each stage of the encoder, cluster it with simple k-means, and visualize the stage-wise clustering to qualitatively  understand the evolving patterns, if any. We empirically choose $k$ equal to $10$. We find roughly $5-6$ of these 10 clusters to be visually explainable, the rest possibly contains either more intricate information or are an issue related to the curse of dimensionality \cite{prml-bishop}.

Figure \ref{fig:layerwise-clustering} shows an example. In this figure, the input and label features are extracted from raw input and label (i.e. dense future fusion maps) BEV maps as input to the frozen encoder. On the other hand, for the output row, we first pass the raw input through our whole encoder-decoder architecture to get the dense reconstruction, and then extract the stage-wise features from these reconstructed maps. The first stage clusters distinguish among the trail (blue), low grass (red), medium-height vegetation (yellow), high bushes (orange), puddles (cyan), and muds (indigo). Deeper stages progressively merge these detailed information in the spatial neighborhoods. 

Notably, for the input features, within the same early stage cluster, such as trail (stage 1 blue), the model separate the noisy and uncertain regions from their correct counterparts in the deeper layers -- i.e. stage $2$ (yellow/blue), stage $3$ (blue/yellow), stage $4$ (red/orange), and stage $5$ (blue/indigo). These uncertain clusters are absent in both label and our reconstructed map features illustrating the advantage of employing reconstructure maps for further downstream tasks, as described next.

\subsection{Evaluation on the Downstream Task}

\noindent \textbf{Costmap estimation:} Following HDIF \cite{hdif}, we first collect the ground truth cost by associating the estimated BEV pixel locations under the tires with the bandpower of the vertical acceleration and shock travel sensor information. 
This highly sparse, speed-conditioned traversability labels makes it impossible to train a standard model for dense costmap prediction from scratch. As noted in Section \ref{subsec:cluster}, the first stage features form a tentative cluster for the traversable terrain (Figure \ref{fig:layerwise-clustering}) with the same spatial high-resolution as the input BEV maps. Thus, we employ the pretrained first stage features from our frozen DBFF model to train a simple 3-layer pixel-wise MLP (Linear $\rightarrow$ BN $\rightarrow$ ReLU) for dense cost prediction (Figure \ref{fig:sample-results}). Here, for raw input and label BEV maps, we extract the first stage DBFF features, whereas for the predicted maps, we feed the BEV prediction again to extract the first stage features from the reconstructed maps.

Table \ref{tab:costmap} shows the speed-stratified comparison of the raw input features with the predicted features (followed by reconstruction) for vanilla and Bayesian counterparts. The dense labels for this comparison is generated from the dense cost from the future fusion label maps (Figure \ref{fig:sample-results}, row 8). The DBFF model performs better across the speed range.

\vspace{0.2cm}

%% file: sections/conclusion.tex
\vspace*{-0.3cm}
\section{Conclusion and Future Work}
\label{sec:conclusion}

\noindent In this paper, we propose a self-supervised, high-resolution, scalable map completion framework for off-road navigation fusing the future information from the complete trajectory. 
To effectively deal with the variable noise and sparsity, we couple the Bayesian mechanism \cite{prbook} with SOTA generative loss formulations on top of the efficient convolutional backbone. 
We demonstrate the efficacy of our framework from both a direct 
generative modeling perspective and through the lens of a
downstream task. 

With the advent of faster computes, we plan to increase the mapping range with more data and employ our dense map predictions to assist in diverse downstream tasks. Finally, we hope our domain-agnostic framework
will encourage the robotics community in general to pursue 
similar self-supervised dense mapping protocol for large-scale 
pretraining in the foreseeable future.
